\documentclass{article}
\usepackage{spconf}
\usepackage{amsmath,graphicx}
\graphicspath{{./img/}}

\usepackage{color}
\usepackage{amsmath}

\usepackage{multirow}
\usepackage{multicol}
\usepackage[utf8]{inputenc}
\usepackage{caption}
\usepackage{afterpage}

\newcommand{\var}[1]{{\operatorname{#1}}}

\title{Cascade Decoder: A Universal Decoding Method for Biomedical Image Segmentation}
\name{Peixian Liang$^{1\star}$, Jianxu Chen$^{2\star}$, Hao Zheng$^{1}$, Lin Yang$^{1}$, Yizhe Zhang$^{1}$ and Danny Z. Chen$^{1}$\thanks{$^{\star}$ These authors contributed equally to this work.}}

\address{$^{1}$ Department of Computer Science and Engineering, University of Notre Dame, Indiana, USA \\
$^{2}$ Allen Institute for Cell Science, Seattle, Washington, USA}

\begin{document}
\ninept
\maketitle

\begin{abstract}
The Encoder-Decoder architecture is a main stream deep learning model for biomedical image segmentation. The encoder fully compresses the input and generates encoded features, and the decoder then produces dense predictions using encoded features. However, decoders are still under-explored in such architectures. In this paper, we comprehensively study the state-of-the-art Encoder-Decoder architectures, and propose a new universal decoder, called {\it cascade decoder}, to improve semantic segmentation accuracy. Our cascade decoder can be embedded into existing networks and trained altogether in an end-to-end fashion. The cascade decoder structure aims to conduct more effective decoding of hierarchically encoded features and is more compatible with common encoders than the known decoders. We replace the decoders of state-of-the-art models with our cascade decoder for several challenging biomedical image segmentation tasks, and the considerable improvements achieved demonstrate the efficacy of our new decoding method. 

\end{abstract}

\section{Introduction}

Image segmentation is a fundamental problem in biomedical image analysis. Recently, deep learning has significantly improved image segmentation accuracy on many problems. Most state-of-the-art deep learning segmentation models are based on the Encoder-Decoder architecture.
The ``encoder" is a typical deep convolutional network 
%(e.g., VGG \cite{simonyan2014very}) 
to encode hierarchical information into feature maps, while the ``decoder" aims to make effective dense predictions using encoded features. The essential spirit of the Encoder-Decoder architecture is to first interpret the images and then predict the segmentation of the images based on the interpretation. 

Many advanced deep learning techniques (e.g., residual learning~\cite{he2016deep})
%and Inception mechanism \cite{szegedy2016rethinking}) 
can be embedded into encoders to better distill information from images. However, the structure of decoders is still under-explored. In the literature, the decoders used in state-of-the-art Encoder-Decoder architectures are mainly of three prototypes: model-wise, scale-wise, and layer-wise decoders (see Section \ref{related} for detailed reviews and discussions). Each such prototype was designed with specific assumptions or functionality.

In this paper, we propose a new universal decoder, called {\it cascade decoder}, which automatically learns the types of rich hierarchical features that are crucial for resolving ambiguity in semantic segmentation. It deeply and effectively consolidates contextual information encoded at different levels, and implicitly embeds deep supervision for efficient training. Our new decoder is ``universal" in the sense that it works well with different types of encoders.  With the term ``cascade", we emphasize that our proposed decoder leverages multi-scale information from coarse to fine. It is important to state that our cascade decoder is a prototype decoding structure. The same idea can be applied to both 2D and 3D Encoder-Decoder architectures, with different specifications of encoders. %The cascade decoder can be easily implemented as it behaves very stably over a wide range of encoders.
With a dense structural encoder, the cascade decoder yields $3.6\%$ dice improvement over the original scale-wise decoder on the 3D HVSMR dataset \cite{pace2015interactive}. With a residual structural encoder, the cascade decoder has $9\%$ dice improvement over the original scale-wise decoder on the 3D pancreas dataset \cite{roth2015deeporgan}. 
Its effectiveness on various datasets demonstrates that our cascade decoder is a competitive alternative to existing decoders that have been dominant for image segmentation tasks.

%Note that in this work, we focus on the study of backbone networks in deep learning models. Other extensions, such as multi-channel networks \cite{xu2017gland} or multi-scale patches \cite{liu2016multi}, can be naturally combined to further boost the segmentation performance.

\section{Related Work}
\label{related}

In this section, we summarize the decoding structures in state-of-the-art Encoder-Decoder architectures for biomedical image segmentation as three prototype decoders. As their counterpart in the Encoder-Decoder architectures, three prototype encoders are also reviewed. 

\begin{figure}[t]
\centering
\includegraphics[width = 3.3in, height = 0.8 in]{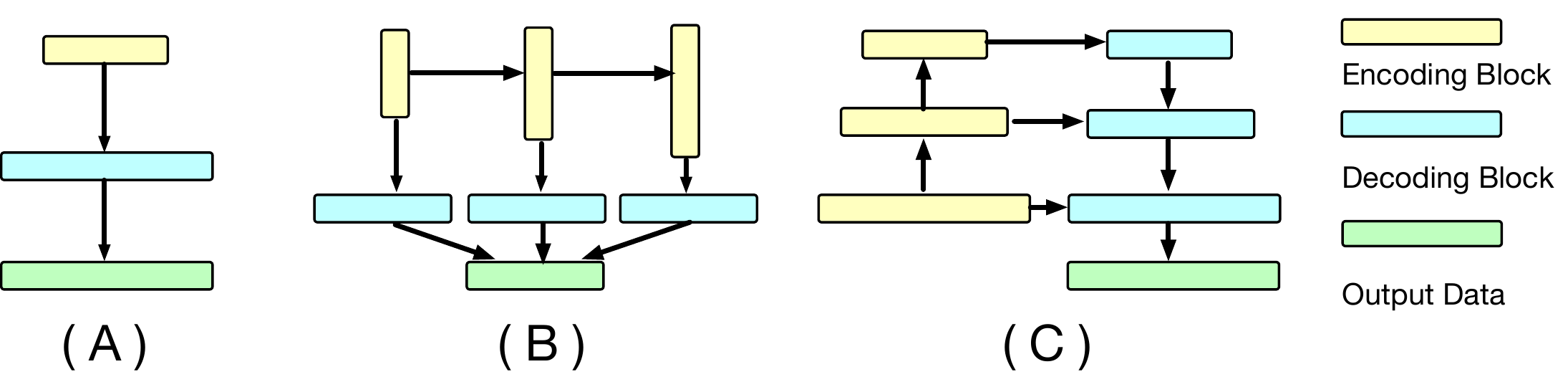}
\caption{Illustrating different prototype decoders: (A) model-wise architecture; (B) scale-wise architecture; (C) layer-wise architecture.}
\label{Decoder}
%\vspace{-0.5cm}
\end{figure}

\begin{figure}
\captionsetup{font=footnotesize}
\centering
\includegraphics[width = 3.3in, height = 1.2in]{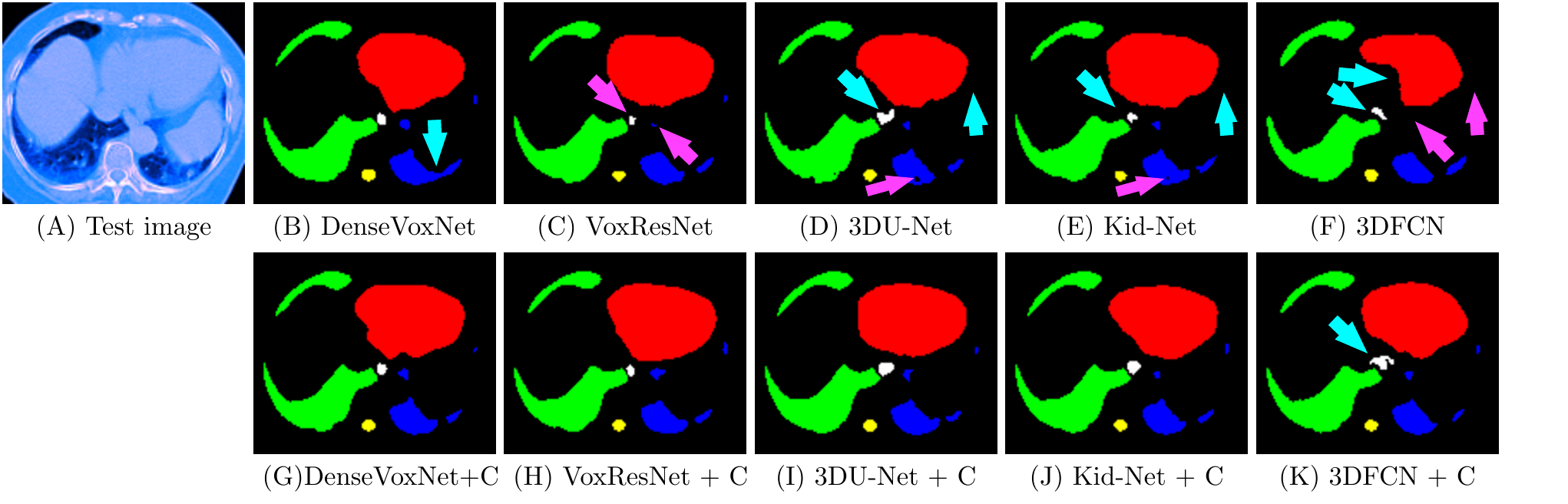}
\caption{\textbf{Qualitative comparison of segmentation results of different models on the 3D AAPM dataset} (red: heart; green: right lung; blue: left lung; yellow: spinal cord; white: esophagus).  The 2nd row shows the results of our method. The ground truth is not supplied (it is held by the challenge organizers). We mark some possible errors by arrows; in particular, purple masks indicate the disadvantages of the model-wise, scale-wise, and layer-wise decoders, as discussed in Section \ref{related}. It shows that our method can achieve better performance (better viewed in color and zoom in).}
\label{result}
\end{figure}

\noindent
\textbf{Decoder 1: Model-wise Decoding}. A typical model-wise decoding structure is shown in Fig.~\ref{Decoder}(A). Its core idea is to treat the whole encoder as a ``blackbox" model, assuming that all the learned information is encoded into the output of the last layer of the encoder network. We call it ``model-wise" because the decoder takes the ``model-wise" output of the encoder model as the single input for decoding. Such structure has been widely used for segmentation tasks ~\cite{chen20163d}. 
%such as for 3D volumetric segmentation \cite{chen20163d} and 2D segmentation \cite{long2015fully,wang2017understanding}.
Model-wise decoders focus more on the semantic context and may have trouble with segmenting fine details. We will show in Section ~\ref{exp} that, for example, such decoders perform poorly in segmenting thin tissues (e.g., see the purple arrows in Fig.~\ref{result}(F)). 

\noindent
\textbf{Decoder 2: Scale-wise Decoding}. The key concept of scale-wise decoding is to 
perform decoding at each scale of the encoder independently (i.e., with different feature map resolutions) using a network with parallel streams, as illustrated in Fig.~\ref{Decoder}(B). Information decoded from different scales is fused to produce the final prediction. Many known models employ scale-wise decoding ~\cite{chen2016voxresnet}. 
%for %3D segmentation \cite{dou20163d, chen2016voxresnet} and 2D segmentation \cite{hariharan2015hypercolumns,chen2016dcan}. segmentation ~\cite{chen2016voxresnet}. 
For this prototype, even though it produces multiple outputs from multiple scales, it does not combine coarse, higher scale information with fine, lower scale information. With such information combination, %for example, 
coarse information could benefit finer level decoding. Sample results will be shown in Section ~\ref{exp} (e.g., see the purple arrows in Fig.~\ref{result}(C)).
\noindent
\textbf{Decoder 3: Layer-wise Decoding}. The decoder design can also be guided by the encoder, layer by layer. On one hand, the structure of a layer in the decoder can be a copy of the corresponding layer in the encoder, except that resolution-reducing operations (e.g., pooling) are replaced by resolution-increasing operations (e.g., deconvolution). On the other hand, the encoded feature map from one layer of the encoder can also be re-used in the corresponding layer of the decoder. As illustrated in Fig.~\ref{Decoder}(C), we call such decoders ``layer-wise decoding". Such structures have been widely used in biomedical image segmentation %\cite{ronneberger2015u,chen2017neuron,cciccek20163d,segnet}.
\cite{ronneberger2015u,chen2017neuron, cciccek20163d}.
While this topology combines coarse information with fine information, %for per-pixel prediction, 
it has only one output prediction. In this way, the information on finer details may interfere with the high-level information. 
%and cause less effective predictions. %In Kid-Net \cite{taha2018kid}, the decoder in such an architecture generates multiple side outputs, but the side outputs are resized to the original scale directly and the final prediction is made by na\"ive average combination. Thus, the above problem still remains. 
Some examples of this kind will be shown in Section \ref{exp} (e.g., see the purple arrows in Fig.~\ref{result}(D)-(E)).

\noindent
\textbf{Representative Encoders.} (1) Linear structure: The idea of this structure is to chain convolutions and pooling operations for feature extraction, such as in 
%VGG ~\cite{simonyan2014very}.
~\cite{chen20163d}.
(2) Residual structure:
In such structure, the residual path element-wisely adds the input features to the output of the same block, making it a residual unit ~\cite{he2016deep}. 
This structure has been developed into various architectures ~\cite{chen2016voxresnet, liang2018new}.
(3) Dense structure: The dense structure uses a densely connected path to concatenate the input features with the output features, allowing each layer to utilize raw information from all the previous layers ~\cite{huang2017densely}. 
This structure has been widely used in FCN models~\cite{yu2017automatic, jegou2017one}.

\section{Methodology}

In this section, we present the formulation of our proposed decoder, which is compatible with different encoders and able to fuse hierarchical information deeply and effectively. Instead of being a specific model, this decoder is a general prototype that can be embedded into existing models with different specifications of encoders.
\\

\noindent
\textbf{Cascade Decoder}

\begin{figure}[t]
\captionsetup{font=footnotesize}
\centering
\includegraphics[width = 3.35 in, height = 1.6 in]{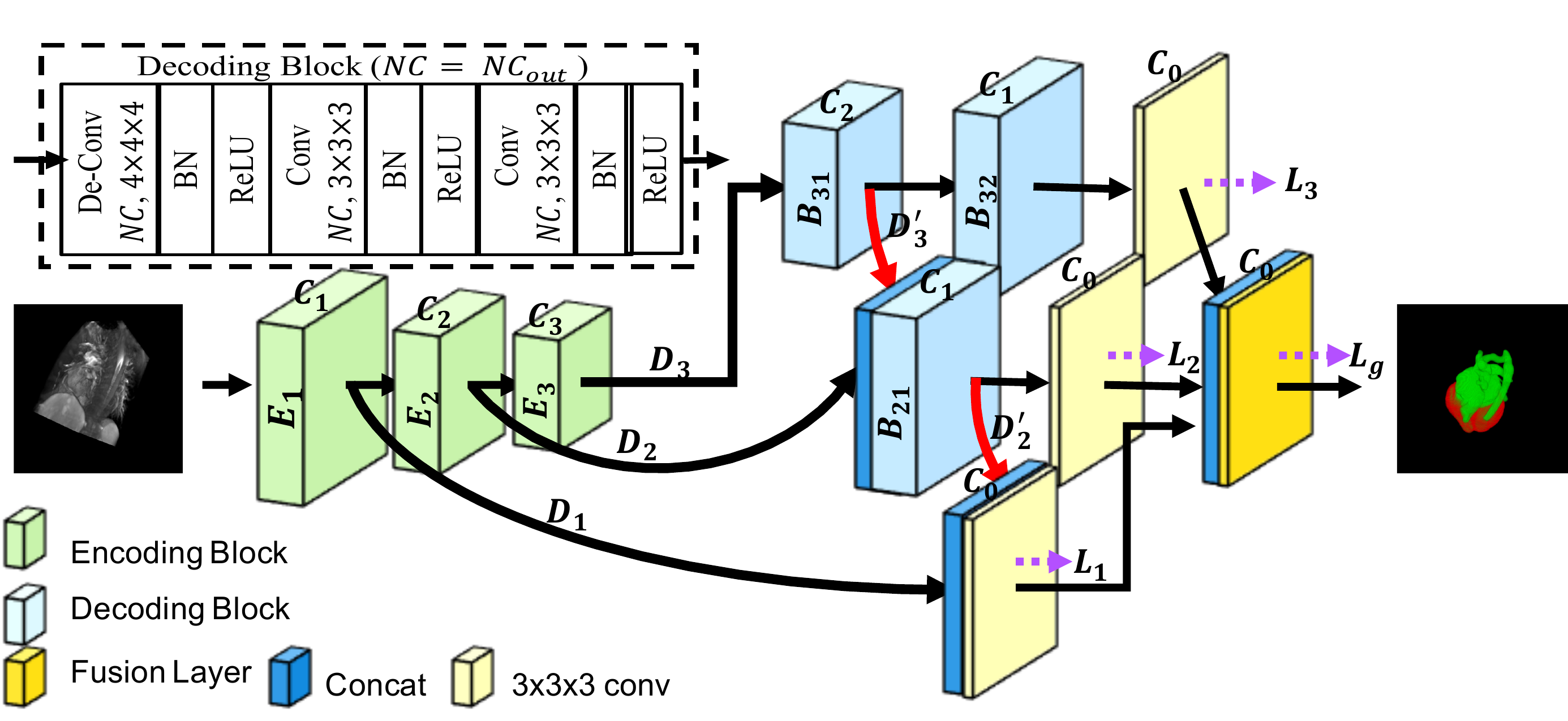}
\caption{\textbf{The macro-architecture of our universal cascade decoder.} For $n$ specific encoding blocks $E_{1}$, $E_{2}$, $\ldots,$ $E_{n}$ ($n=3$ in this figure), there are $n$ corresponding decoding branches $D_{1}$, $D_{2}$, $\ldots,$ $D_{n}$.
$B_{ij}$ denotes the decoding block $j$ at decoding branch $i$. Side-branch $D'_{i}$ (red arrow) concatenates the output features from $B_{i1}$ with the output features from $E_{i-1}$ and feeds to $B_{(i-1)1}$. $L_{i}$ is the auxiliary loss at branch $i$, and $L_{g}$ is the global loss. $C_{i}$ above each block is the number of output channels. %($C_{0}$ is equal to the number of classes). 
At the end, the fusion layer is performed to generate the probability map as the result.}
\label{Framework}
\end{figure}

\noindent
Fig.~\ref{Framework} shows an overview of our proposed decoder (for 3D images). It employs a cascade branching structure to consolidate hierarchical feature maps from different encoding blocks (i.e., different scales). Suppose our decoder is working with an encoder with $k$ encoding blocks ($k$ different scales). For each encoding block $E_{i}$, $i=2,\ldots,k$, there is a commensurate decoding branch $D_{i}$ consisting of a sequence of decoding blocks ${B_{i1}, B_{i2},\ldots, B_{i(i-1)}}$. $D_{i}$ gradually decodes features from low resolution to high resolution and makes prediction at the end of the branch. For encoding block $E_{1}$, it has a decoding branch $D_{1}$ with a $3\times3\times3$ convolution for 3D ($3\times3$ for 2D) to produce prediction. Finally, predictions from different branches are concatenated together and go through the fusion layer to produce the final result. Except for $D_{1}$, each decoding branch $D_{i}$ generates a \textbf{side-branch} $D'_{i}$ (shown by red arrows in Fig.~\ref{Framework}) after $B_{i1}$. The purpose of side-branch $D'_{i}$ is to propagate the information from a coarser scale to guide the decoding at a finer scale. Specifically, the output of $B_{i1}$ and the output of $E_{i-1}$ are concatenated and fed into $B_{(i-1)1}$ for decoding. In addition, in order to alleviate the problem of vanishing gradients and impose direct supervision on each branch, the auxiliary loss functions are applied to the prediction from each branch $D_{i}$, besides the final global loss. Furthermore, the auxiliary loss can serve as an additional constraint to the learning process, which can improve the discriminability and robustness of the learned features in intermediate layers \cite{lee2015deeply}.

 In our cascade decoder, a key basic element is the \textbf{decoding block} (DB) $B_{ij}$, as shown in Fig.~\ref{Framework}. In each decoding block for 3D segmentation models, a $4\times4\times4$ deconvolution with stride 2 is applied for upsampling, followed by batch normalization (BN) and ReLU activation. Then two successive ``$3\times3\times3$ convolution $+BN+ReLU$" are applied. In 2D segmentation models, we use $4 \times 4$ deconvolution and $3 \times 3$ convolutions instead.

 Since our cascade decoder has distinct branches and works in different scales to produce diverse predictions, for different objects, there is a better chance that one of the predictions would give correct results. In order to obtain an optimal prediction, we need a \textbf{fusion layer} (as shown in Fig.~\ref{Framework}), which is capable of learning robust visual features for jointly utilizing the diverse prediction results. Thus, we employ a fusion layer with $1\times1\times1$ convolution for 3D ($1\times1$ for 2D) to fuse outputs and produce the final prediction.

The cascade structure of our proposed decoder has three main advantages. First, the side-branch ($D'_{i}$) feeds the output of a decoding block $B_{i1}$  into a decoding block $B_{(i-1)1}$, which allows the decoding blocks to use semantic feature maps from the previous decoding block (higher-scale) to correct any potential errors introduced by lower-scale. Second, each decoding branch can completely and independently perform decoding from the corresponding encoding scale and make its prediction using a sequence of decoding blocks. Third, the fusion layer can help fuse outputs from different branches and improve the segmentation quality.
Our experiments show that these key components play a big role in improving the segmentation performance (see Table~\ref{tab:Abl}).

\begin{table}[ht!]
\captionsetup{font=footnotesize}
\centering
\caption{ \textbf{3D and 2D FCN models used in the experiments and their prototypes.} The Encoders and Decoders in this table correspond to the prototype encoders and decoders discussed in Section~\ref{related}. The numbers in the parentheses represent the numbers of encoding blocks.}
\label{tab:3DFCN}
%\tiny
%\scriptsize
\footnotesize
%\renewcommand\arraystretch{1.4}
% adjust the height of cells of table
\setlength{\tabcolsep}{0.3mm}
{
\begin{tabular}{ c|c c c c c c}

\hline
  \multirow{ 5}{*}{3D } & & DenseVoxNet &  VoxResNet &  3D U-Net & Kid-Net & 3D FCN \\
  
  & & (DVN) \cite{yu2017automatic}   &  (VRN) \cite{chen2016voxresnet}    &  (U-Net) \cite{cciccek20163d} & (KD) \cite{taha2018kid} & (FCN) \cite{chen20163d} \\ \cline{2-7}

 &Encoder &  Dense (2)  &  Residual (4)  &  Linear (4) & Linear (5) & Linear (3)  \\ 

 & Decoder & Scale-wise &  Scale-wise   &  Layer-wise & Layer-wise & Model-wise   \\ 
 
\hline
\hline

  \multirow{ 4}{*}{2D } & & U-Net  &  ResNet   &  DenseNet & & \\ 
  
  & & (UNet)~\cite{ronneberger2015u}  &  (RN) \cite{chen2016voxresnet}    &  (DN)~\cite{jegou2017one} & & \\ \cline{2-7}

& Encoder &  Linear (5)  &  Residual (4)  &  Dense (4) & &\\ 

& Decoder & Layer-wise &  Scale-wise   &  Layer-wise & &\\ 

\hline

\end{tabular}
}
\end{table}

\begin{table}[ht!]
\captionsetup{font=footnotesize}
\centering
\caption[dd]{ \textbf{Segmentation result comparison on the 3D HVSMR dataset\protect\footnotemark}.}

\label{tab:HVSMR}
%\tiny
%\scriptsize
\footnotesize
%\renewcommand\arraystretch{1.4}
% adjust the height of cells of table
\setlength{\tabcolsep}{0.3mm}
{
\begin{tabular}{c c c c c c c}

\hline
\multirow{ 2}{*}{ Model }  &   \multicolumn{3}{c}{Myocardium} & \multicolumn{3}{c}{Blood Pool}  
\\  \cline{2-7}
 & Dice   &  ADB[mm]    &  HD[mm]   & Dice   &  ADB[mm]    &  HD[mm] \\ \hline

DVN \cite{yu2017automatic}   &   $ 0.792 $  &  $ 0.815 $  &  $ 4.701 $  & $ \textbf{0.936} $  &  $ 0.871 $  &  $ 8.265 $     \\ 

DVN + C (ours)  &    $ \textbf{0.828} $  &  $ \textbf{0.681} $   &  $ \textbf{3.686} $  &  $ \textbf{0.936} $  & $ \textbf{0.793} $ &  $ \textbf{6.719} $  \\ 
\hline

 VRN \cite{chen2016voxresnet} &  $ 0.789 $  &  $ 0.814 $   &  $ 4.394 $  &  $ \textbf{0.935} $  & $ 0.868 $ &  $ 8.022 $  \\ 

 VRN + C (ours) &  $ \textbf{0.800} $  &  $ \textbf{0.780} $   &  $ \textbf{4.254} $  &  $ \textbf{0.935} $  & $ \textbf{0.799} $ &  $ \textbf{5.969} $  \\ 
 \hline

U-Net \cite{cciccek20163d} &  $ 0.762 $  &  $ 0.943 $   &  $ 5.618 $  &  $ 0.932 $  & $ 0.826 $ &  $ 7.015 $   \\ 
 
U-Net + C (ours) &  $ \textbf{0.801} $  &  $ \textbf{0.767} $   &  $ \textbf{4.094} $  &  $ \textbf{0.934} $  & $ \textbf{0.806} $ &  $ \textbf{6.828} $  \\
 \hline
 
KN \cite{taha2018kid}
  &  $ 0.785 $  &  $ 0.935 $   &  $ 6.061 $  &  $ 0.933 $  & $ 0.816 $ &  $ 6.791 $   \\ 
 
KN + C (ours) &  $ \textbf{0.816} $  &  $ \textbf{0.720} $   &  $ \textbf{4.142} $  &  $ \textbf{0.938} $  & $ \textbf{0.787} $ &  $ \textbf{6.694} $  \\

 \hline  
 FCN \cite{chen20163d} &  $ 0.782 $  &  $ 0.883 $   &  $ 5.006 $  &  $ 0.933 $  & $ 0.898 $ &  $ \textbf{8.106} $   \\ 
 
FCN + C (ours) &  $ \textbf{0.821} $  &  $ \textbf{0.736} $   &  $ \textbf{4.283} $  &  $ \textbf{0.934} $  & $ \textbf{0.878} $ &  $ 8.194 $  \\

\hline
\end{tabular}
}
\end{table}

\begin{table*}[ht!]
\captionsetup{font=footnotesize}
\centering
\caption{ \textbf{Segmentation result comparison on the 3D AAPM dataset.} }%DVN = DenseVoxNet, VRN = VoxResNet, UNet = 3D U-Net, KN = Kid-Net, FCN = 3D FCN, and ``+ C" means replacing the decoder in the corresponding model with the cascade decoder.}
\label{tab:AAPM}
%\tiny
%\scriptsize
\footnotesize
%\renewcommand\arraystretch{1.4}
% adjust the height of cells of table
\setlength{\tabcolsep}{0.3mm}
{
\begin{tabular}{c c c c | c c c | c c c | c c c | c c c}

\hline
\multirow{ 2}{*}{ Model }  &   \multicolumn{3}{c}{Esophagus} & \multicolumn{3}{c}{SpinalCord}  &   \multicolumn{3}{c}{Lung $\_$ R} &   \multicolumn{3}{c}{Lung $\_$ L} &   \multicolumn{3}{c}{Heart}
\\  \cline{2-16}
 & Dice   &  ADB[mm]    &  HD[mm]   & Dice   &  ADB[mm]    &  HD[mm] & Dice   &  ADB[mm]    &  HD[mm] & Dice   &  ADB[mm]    &  HD[mm] & Dice   &  ADB[mm]    &  HD[mm]\\ \hline

DVN \cite{yu2017automatic}   &   $ 0.676 $  &  $ 2.227 $  &  $ 7.748 $  & $ 0.851 $  &  $ 0.867 $  &  $ 2.298 $  &  $ 0.960 $  &  $ 1.212 $  &  $ 3.938 $  &   $ \textbf{0.960} $  &  $ 1.295 $  &  $ 4.100 $ &   $ \textbf{0.917} $  &  $ \textbf{2.418} $  &  $ \textbf{6.781} $\\ 

DVN + C (ours)  &    $  \textbf{0.683} $  &  $ \textbf{1.978} $   &  $ \textbf{6.458} $  &  $ \textbf{0.864} $  & $ \textbf{0.799} $ &  $ \textbf{2.136} $ &   $ \textbf{0.963} $  &  $ \textbf{1.211} $  &  $ \textbf{3.937} $ &   $ \textbf{0.960} $  &  $ \textbf{1.063} $  &  $ \textbf{3.368} $ &   $ 0.914 $  &  $ 2.553 $  &  $ 7.430 $ \\ 
\hline

 VRN \cite{chen2016voxresnet} &  $ 0.658 $  &  $ 2.398 $   &  $ 8.194 $  &  $ 0.809 $  & $ 1.126 $ &  $ \textbf{3.039} $ &   $ 0.956 $  &  $ 1.397 $  &  $ 4.140 $ &   $ 0.953 $  &  $ 1.204 $  &  $ 3.679 $ &   $ 0.895 $  &  $ 3.007 $  &  $ 7.836 $ \\ 

 VRN + C (ours) &  $ \textbf{0.676} $  &  $ \textbf{2.151} $   &  $ \textbf{7.294} $  &  $ \textbf{0.839} $  & $ \textbf{1.052} $ &  $ 3.173 $ &   $ \textbf{0.959} $  &  $ \textbf{1.322} $  &  $ \textbf{4.012} $ &   $ \textbf{0.957} $  &  $ \textbf{1.127} $  &  $ \textbf{3.499} $ &   $ \textbf{0.911} $  &  $ \textbf{2.607} $  &  $ \textbf{7.480} $ \\ 
 \hline

U-Net \cite{cciccek20163d} &  $ \textbf{0.708} $  &  $ 1.937 $   &  $ 6.704 $  &  $ 0.854 $  & $ 0.837 $ &  $ 2.204 $ &   $ 0.961 $  &  $ 1.234 $  &  $ \textbf{3.873} $ &   $ 0.960 $  &  $ 1.050 $  &  $ 3.311 $ &   $ 0.918 $  &  $ 2.416 $  &  $ \textbf{6.906} $  \\ 
 
U-Net + C (ours) &  $ 0.703 $  &  $ \textbf{1.899} $   &  $ \textbf{6.226} $  &  $ \textbf{0.878} $  & $ \textbf{0.719} $ &  $ \textbf{2.038} $  &   $ \textbf{0.965} $  &  $ \textbf{1.170} $  &  $ 3.947 $ &   $ \textbf{0.962} $  &  $ \textbf{1.014} $  &  $ \textbf{3.193} $ &   $ \textbf{0.919} $  &  $ \textbf{2.392} $  &  $ 7.123 $\\
 \hline  
 
 KN \cite{taha2018kid} &  $ 0.652 $  &  $ 2.413 $   &  $ 8.448 $  &  $ 0.827 $  & $ 0.987 $ &  $ 2.542 $ &   $ 0.956 $  &  $ 1.372 $  &  $ \textbf{3.946} $ &   $ \textbf{0.956} $  &  $ \textbf{1.125} $  &  $ \textbf{3.421} $ &   $ 0.903 $  &  $ 2.870 $  &  $ 8.034 $  \\ 
 
KN + C (ours) &  $ \textbf{0.653} $  &  $ \textbf{2.264} $   &  $ \textbf{7.308} $  &  $ \textbf{0.871} $  & $ \textbf{0.766} $ &  $ \textbf{2.115} $  &   $ \textbf{0.961} $  &  $ \textbf{1.128} $  &  $ 4.078 $ &   $ \textbf{0.956} $  &  $ 1.138 $  &  $ 3.515 $ &   $ \textbf{0.914} $  &  $ \textbf{2.532} $  &  $ \textbf{7.077} $\\
 \hline  
 
 FCN \cite{chen20163d}&  $ 0.561 $  &  $ 3.365 $   &  $ 9.981 $  &  $ 0.757 $  & $ 1.917 $ &  $ 5.876 $ &   $ 0.944 $  &  $ 2.310 $  &  $ 8.306 $ &   $ 0.936 $  &  $ 1.767 $  &  $ 6.716 $ &   $ 0.884 $  &  $ 3.532 $  &  $ 9.985 $  \\ 
 
FCN + C (ours) &  $ \textbf{0.631} $  &  $ \textbf{2.447} $   &  $ \textbf{7.345} $  &  $ \textbf{0.823} $  & $ \textbf{1.043} $ &  $ \textbf{3.069} $  &   $ \textbf{0.957} $  &  $ \textbf{1.158} $  &  $ \textbf{3.591} $ &   $ \textbf{0.955} $  &  $ \textbf{1.408} $  &  $ \textbf{4.302} $ &   $ \textbf{0.901} $  &  $ \textbf{3.088} $  &  $ \textbf{9.058} $\\
 \hline  
 
\end{tabular}
}
\end{table*}

%\begin{table}[ht!]
%\captionsetup{font=footnotesize}
%\centering
%\caption{ \textbf{Segmentation result comparison on the 3D pancreas dataset.}} %DVN = DenseVoxNet, VRN = VoxResNet, UNet = 3D U-Net, KN = Kid-Net, FCN = 3D FCN, and ``+ C" means replacing the decoder in the corresponding model with the cascade decoder.}\

%\label{tab:NIH1}
%%\tiny
%%\scriptsize
%\footnotesize
%%\renewcommand\arraystretch{1.4}
%% adjust the height of cells of table
%\setlength{\tabcolsep}{0.3mm}
%{
%  \begin{tabular}{c c c c }

%\hline
% Model & Dice   &  ADB[mm]    &  HD[mm]   \\ \hline

%DenseVoxNet \cite{yu2017automatic} &  $ 0.820 $  &  $ 0.474 $   &  $ \textbf{15.355} $  \\ 

% DVN + C (ours) &  $ \textbf{0.823} $  &  $ \textbf{0.472} $   &  $ 15.444 $   \\ 
% \hline
 
% VoxResNet \cite{chen2016voxresnet} &  $ 0.752 $  &  $ 0.815 $   &  $ 21.872 $  \\ 

%VRN + C (ours) &  $ \textbf{0.841} $  &  $ \textbf{0.376} $   &  $ \textbf{16.685} $   \\ 
% \hline

%3D U-Net \cite{cciccek20163d} &  $ 0.835 $  &  $ 0.475 $   &  $ 15.068 $    \\

%UNet + C (ours) &  $ \textbf{0.841} $  &  $ \textbf{0.373} $   &  $ \textbf{14.491} $   \\ 
% \hline
 
%Kid-Net \cite{taha2018kid} &  $ 0.823 $  &  $ 0.476 $   &  $ 17.324 $    \\

%KD + C (ours) &  $ \textbf{0.843} $  &  $ \textbf{0.362} $   &  $ \textbf{14.773} $   \\ 
% \hline
 
% 3D FCN \cite{chen20163d}&  $ 0.804 $  &  $ 0.675 $   &  $ 27.891 $    \\

%FCN + C (ours) &  $ \textbf{0.831} $  &  $ \textbf{0.503} $   &  $ \textbf{17.960} $   \\ 
% \hline
 
%\end{tabular}
%}
%\end{table}

\begin{table}[ht!]
\captionsetup{font=footnotesize}
\centering
\caption{ \textbf{Segmentation result comparison on the 3D pancreas dataset.}} %DVN = DenseVoxNet, VRN = VoxResNet, UNet = 3D U-Net, KN = Kid-Net, FCN = 3D FCN, and ``+ C" means replacing the decoder in the corresponding model with the cascade decoder.}\

\label{tab:NIH}
%%\tiny
%%\scriptsize
\footnotesize
%%\renewcommand\arraystretch{1.4}
%% adjust the height of cells of table
\setlength{\tabcolsep}{0.3mm}
{
  \begin{tabular}{c |c c | c c | c c }

\hline

& DVN &  DVN + C & VRN  & VRN + C & U-Net & U-Net + C   \\

& \cite{yu2017automatic} & (ours) & \cite{chen2016voxresnet}& (ours) & \cite{cciccek20163d} & (ours)  \\ \hline

 Dice & 0.820 &  $ \textbf{0.823} $   &  $ 0.752 $ & $\textbf{0.841}$ & $0.835$ & $\textbf{0.841}$ \\ 
 \hline
 
 ADB[mm] &  $ 0.474 $  &  $\textbf{0.472}$   &  $ 0.815 $ & $\textbf{0.376}$ & $0.475$ & $\textbf{0.373}$  \\ \hline

HD[mm] &  $ \textbf{15.355} $  &  $ 15.444 $  &  $21.872 $ & $\textbf{16.685}$ & $15.068 $ & $\textbf{14.491}$   \\ 
 \hline

\end{tabular}
}
\end{table}

\begin{table}[ht!]
\captionsetup{font=footnotesize}
\centering
\caption{ \textbf{Segmentation result comparison on the 2D Fungus dataset.}} %RN = ResNet, DN = DenseNet %and ``+ C" means replacing the decoder in the corresponding model with the cascade decoder.}
\label{tab:Fungus}
%\tiny
%\scriptsize
\footnotesize
%\renewcommand\arraystretch{1.4}
% adjust the height of cells of table
\setlength{\tabcolsep}{0.5mm}
{
\begin{tabular}{c |c c | c c | c c}

\hline
\multirow{ 2}{*}{} & UNet & UNet + C  & RN  & RN + C  & DN  &  DN + C \\ 

&\cite{ronneberger2015u} & (ours) & \cite{liang2018new}  & (ours) & \cite{jegou2017one} & (ours) \\ \hline

IoU &  $ 0.941 $  &  $ \textbf{0.950} $ & $0.940$ & $\textbf{0.948}$ & $0.930$ & $\textbf{0.935}$ \\ 

 F1 &   $0.944$  &  $ \textbf{0.953} $ & $0.945$ & $\textbf{0.950}$ & $0.932$ & $\textbf{0.938}$ \\ 
 \hline

\end{tabular}
}
\end{table}

\begin{table}[ht!]
\captionsetup{font=footnotesize}
\centering
\caption{ \textbf{Segmentation ablation results on the 3D HVSMR dataset.}} %DVN = DenseVoxNet, VRN = VoxResNet, DB = decoding block, and ``+ C" means replacing the decoder in the corresponding model with the cascade decoder.}
\label{tab:Abl}
%\tiny
%\scriptsize
\footnotesize
%\renewcommand\arraystretch{1.4}
% adjust the height of cells of table
\setlength{\tabcolsep}{0.8mm}
{
\begin{tabular}{c c c c c c c}

\hline
\multirow{ 3}{*}{ Model }  &   \multicolumn{3}{c}{Myocardium} & \multicolumn{3}{c}{Blood Pool}  
\\  \cline{2-7}
 & Dice   &  ADB   &  HD  & Dice   &  ADB  &  HD \\ 
 & &  [mm]    &  [mm]   &   &  [mm]    & [mm] \\ \hline
 
DVN + C &    $ \textbf{0.828} $  &  $ \textbf{0.681} $   &  $ \textbf{3.686} $  &  $ \textbf{0.936} $  & $ \textbf{0.793} $ &  $ 6.719 $  \\ 

DVN + C (w/o side-branch)  &  $ 0.820 $  &  $ 0.682 $   &  $ 3.813 $  &  $ \textbf{0.936} $  & $ 0.848 $ &  $ 7.832 $  \\ 

DVN + C (w/o fusion layer)  &  $ 0.801 $  &  $ 0.765 $   &  $ 4.455 $  &  $ 0.935 $  & $ 0.811 $ &  $ \textbf{6.603} $  \\ 
\hline

VRN + C &  $ \textbf{0.800} $  &  $ \textbf{0.780} $   &  $ \textbf{4.254} $  &  $ 0.935 $  & $ \textbf{0.799} $ &  $ \textbf{5.969} $  \\ 

VRN + C (w/o seq. of DBs) &  $0.795$   & $ 0.858 $  &  $ 4.816 $  & $ \textbf{0.936} $  & $ 0.858 $ & $ 8.207 $  \\ 
\hline

%\rule{0in}{1.2em}$^\dag$\scriptsize DBs represents decoding blocks.\\ 

\end{tabular}
}
\end{table}

\section{Evaluations}

\label{exp}
\footnotetext{In the context of this paper, %DVN = DenseVoxNet, VRN = VoxResNet, UNet = 3D U-Net, KN = Kid-Net, FCN = 3D FCN, and 
``+ C" means replacing the decoder in the corresponding model with the cascade decoder.}

To compare with existing decoders working with different encoders, we replace the original decoders in the state-of-the-art deep learning models with our cascade decoder. For the three prototypes of known encoders discussed in Section~\ref{related}, we choose at least one typical representative from each prototype. For each specific encoder, it has a corresponding decoder for comparison with our decoder. The FCN models used in our experiments %and their corresponding prototypes 
are shown in Table~\ref{tab:3DFCN}. 
%These FCN models are commonly used for biomedical image segmentation tasks. %We use two public 3D datasets and one in-house 2D dataset to verify the performance of our method for 2D and 3D segmentation. 

\noindent
\textbf{Data.} We conduct extensive experiments on four datasets. (1) The 3D HVSMR dataset is public data from the %HVSMR 2016
Challenge \cite{pace2015interactive}, aiming to segment the blood pool and ventricular myocardium in 3D cardiovascular MR images. %It has 10 training scans and 10 test scans. 
We employ spatial resampling to $1mm$ isotropically. (2) The 3D thoracic dataset is from the AAPM 2017 Challenge \cite{yang2017thoracic}. This challenge aimed to segment esophagus, spinal cord, left lung, right lung, and heart. %36 3D CT images were provided by the challenge organizers as training data. The test data consist of another 12 3D CT images. 
We rescale each slice to a $\var{3-}$channel $\var{8-}$bit image using three windows of [-1000, 600], [-160, 240], and [-1000, -550] Hounsfield units.
%\noindent
(3) The 3D pancreas segmentation dataset is from publicly available data \cite{roth2015deeporgan}, whose purpose is to segment pancreas in 3D contrast-enhanced abdominal CT volumes. %82 3D volumes are supplied. 
We use 61 volumes as training data and 21 volumes as testing data. 
%We resize the raw volumes with the ratio of 0.5. 
(4) The 2D fungus dataset has in-house EM images for analyzing tubular fungal structures, with 21 images of size $1658 \times 1705$. We use 11 images as training data and 10 images as testing data. 

%For the HVSMR and AAPM datasets, the test ground truth data are held by the challenge organizers for fair comparison. For the 3D segmentation tasks, the results are evaluated using three criteria: (1) Dice coefficient, (2) average surface distance (ADB), and (3) symmetric Hausdorff distance (HD). For the 2D fungus dataset, the results are evaluated by F1 score and mean IoU. 

\noindent
\textbf{Implementation details.} Our FCN models are implemented with TensorFlow~\cite{abadi2016tensorflow}. An NVIDIA Tesla v100 GPU with 32 GB GPU memory is used for both training and testing. The weights of all the models are initialized with a Gaussian distribution. %($\mu = 0$, $\sigma = 0.01$). 
We train our networks using Adam optimization~\cite{kingma2014adam} with a learning rate $5\var{e-}4$. 

%To leverage the limited training data, we train our networks using standard data augmentation strategies (e.g., random rotation with 90, 180, and 270 degrees, and image flipping along the axial planes). For the AAPM dataset, since the left lung and right lung are segmented separately, the rotation and flipping augmentations are not used. We randomly crop patches of size $64 \times 64 \times 64$ to train the 3D models and crop patches of size $192 \times 192$ to train the 2D models. 

\noindent
\textbf{Main Experimental Results} 

\noindent
Table~\ref{tab:HVSMR} shows a quantitative comparison on the HVSMR 2016 Challenge dataset \cite{pace2015interactive}. First, comparing to the baseline, DenseVoxNet, our decoder can boost the performance on all the metrics (``{DVN + C (ours)}", Table~\ref{tab:HVSMR}). Then, for the other models, they can achieve state-of-the-art performance using our decoder.

Table \ref{tab:AAPM} shows a quantitative comparison on the AAPM 2017 Challenge dataset \cite{yang2017thoracic}. %Multi-organ segmentation in CT images is a difficult problem. 
The results show that our decoder can considerably improve the performance on this challenging segmentation task, especially for esophagus and spinal cord. The cascade decoder can increase the esophagus Dice by about $7\%$ and the spinal cord Dice by about $7\%$ on 3D FCN (``{FCN + C (ours)}", Table \ref{tab:AAPM}), and obtain over $4\%$ spinal cord Dice improvement on Kid-Net (``{KN + C (ours)}", Table \ref{tab:AAPM}). Some visual results are given in Fig.~\ref{result}. 

Table~\ref{tab:NIH} shows a quantitative comparison on the 3D pancreas dataset \cite{roth2015deeporgan}. One can see that our decoder can considerably improve the segmentation results, especially for VoxResNet, achieving about $9\%$ Dice improvement. %(probably because the scale-wise decoder used in VoxResNet is not able to decode information from the advanced residual encoder completely).

Table~\ref{tab:Fungus} gives segmentation results on the 2D fungus dataset. This dataset is relatively less challenging than the 3D datasets. 
One can see that our cascade decoder still achieves slightly better performance than the state-of-the-art models on this  segmentation task. 

%In summary, the segmentation results on all 3D/2D datasets demonstrate the effectiveness of our new decoding method. 
%Our cascade decoder can considerably boost the segmentation performance of the state-of-the-art models. 

\noindent
\textbf{Ablation Study}

\noindent
%Here we examine the roles of the key components in our new decoding method for image segmentation. 
Our ablation study uses the 3D HVSMR dataset \cite{pace2015interactive} as an example. In the \textit{side-branch} and \textit{fusion layer} experiments, we use the encoder from DenseVoxNet~\cite{yu2017automatic}, which is the baseline for this dataset. For the \textit{decoding block} experiment, 
%since DenseVoxNet has only two encoding blocks and there is no difference between using consecutive decoding blocks and directly resizing to the original size, 
we use the encoder in VoxResNet~\cite{chen2016voxresnet} instead, which has four encoding blocks. 

\noindent
\textbf{The role of side-branches}. Side-branches are used to combine coarse high-layer information with fine low-layer information. 
% We explore how this design helps in image segmentation. 
We experiment with removing all the side-branches in our decoder. The result is shown in {``DVN + C (w/o side-branch)"} of Table~\ref{tab:Abl}. One can see that the decoder without side-branches performs worse, demonstrating the effectiveness of using side-branches.

\noindent
\textbf{The role of consecutive decoding blocks}.
%A sequence of decoding blocks, $B_{i1}, \ldots,$ $B_{i(i-1)}$, in branch $D_{i}$ can independently perform decoding at the corresponding scale.
In our experiment, the sequence of decoding blocks (DBs) $B_{i2}, \ldots,$ $B_{i(i-1)}$ in $D_{i}$ $(i=3,\ldots,k)$ is removed and a $t \times t \times t$ $(t=2^{i-2}+2)$ deconvolution with a stride $2^{i-2}$ is performed instead, to resize the output from $B_{i1}$ to the raw image size. %$B_{i1}$ is used to upsample the feature map from $E_{i}$ and then propagate to the lower-scale $B_{(i-1)1}$. There is no change in  $D_{1}$ and $D_{2}$, since $D_{1}$ has no decoding block and $D_{2}$ has only one decoding block $B_{21}$. 
The result is given in Table \ref{tab:Abl} (denoted by {``VRN + C (w/o seq. of DBs)"}). One can see that using a sequence of decoding blocks can help achieve better performance, demonstrating the effect of consecutive decoding blocks. 

\noindent
\textbf{The role of fusion layer}. We compare our fusion layer with the implementation of na\"ive average combination (in which the fusion layer is removed and the average of the outputs from different branches is used as the final prediction directly). Table \ref{tab:Abl} ({``DVN + C (w/o fusion layer)"}) shows the result. One can see that average combination is worse than the fusion layer, which demonstrates that the fusion layer can help fuse the side outputs. and thus improve the segmentation performance.

\section{Conclusions}

We proposed a new universal decoder, called cascade decoder, for deep learning based biomedical image segmentation. The main advantage of our cascade decoder is to leverage features from multiple scales efficiently to produce accurate dense predictions. It matches better with encoders than known decoders used in common deep learning models. Qualitative and quantitative experimental results show that our new decoder outperforms state-of-the-art decoders.

\section{Acknowledgement}

This research was supported in part by the National Science Foundation under Grants CNS-1629914, CCF-1640081, and CCF-1617735, and by the Nanoelectronics Research Corporation, a wholly-owned subsidiary of the Semiconductor Research Corporation, through Extremely Energy Efficient Collective Electronics, an SRC-NRI Nanoelectronics Research Initiative under Research Task ID 2698.005. 

%\clearpage



%\bibliographystyle{IEEEtran}


\bibliographystyle{ieeetr}
\bibliography{refs}

\end{document}